\documentclass[conference]{IEEEtran}
\IEEEoverridecommandlockouts

\usepackage{color}
\usepackage{cite}
\usepackage{tabularx}
\usepackage{tikz}
\usepackage{soul}
\usepackage{multirow}
\usepackage{cite}
 
\usepackage{graphicx}
\usepackage{float}
\usepackage{subfigure}
\usepackage{cite} 
\usepackage{amsmath}

\usepackage{cite}
\usepackage{amsmath,amsfonts}
\usepackage{algorithm}
\usepackage{array}
\usepackage{algpseudocode}
\usepackage[caption=false,font=normalsize,labelfont=sf,textfont=sf]{subfig}
\usepackage{textcomp}
\usepackage{stfloats}
\usepackage{url}
\usepackage{verbatim}
\usepackage{graphicx}
\usepackage{subfigure}
\usepackage{cite}
\usepackage{pifont}
\usepackage{makecell}
\usepackage{bm}
\usepackage{bbm}
\usepackage[english]{babel}
\usepackage{amsthm}
\usepackage{subcaption}
\usepackage{color}
\usepackage{subfigure}

\IEEEoverridecommandlockouts

\usepackage{tabularx}
\usepackage{tikz}
\usepackage{soul}
\usepackage{multirow}
 
\usepackage{graphicx}
\usepackage{float}

\begin{document}

\title{Hyperdimensional Computing Empowered Federated Foundation Model over Wireless Networks for Metaverse
}

\author{\IEEEauthorblockN{Yahao Ding\IEEEauthorrefmark{1}, 
Wen Shang\IEEEauthorrefmark{1}, 
Minrui Xu\IEEEauthorrefmark{2},
Zhaohui Yang\IEEEauthorrefmark{3},
Ye Hu\IEEEauthorrefmark{4},
Dusit Niyato\IEEEauthorrefmark{2},
Mohammad Shikh-Bahaei\IEEEauthorrefmark{1}}
\IEEEauthorblockA{\IEEEauthorrefmark{1}King's College London,\IEEEauthorrefmark{2}Nanyang Technological University,\IEEEauthorrefmark{3} Zhejiang University,\IEEEauthorrefmark{4} University of Miami\\
Email: \IEEEauthorrefmark{1}{\{yahao.ding, wen.shang, m.sbahaei\}}@kcl.ac.uk,
\IEEEauthorrefmark{2}{\{minrui001, dniyato\}}@ntu.edu.sg,\\
\IEEEauthorrefmark{3}zhaohuiyang@zju.edu.cn,
\IEEEauthorrefmark{4}yehu@miami.edu}}

\markboth{Journal of \LaTeX\ Class Files,~Vol.~14, No.~8, August~2015}%
{Shell \MakeLowercase{\textit{et al.}}: Bare Demo of IEEEtran.cls for IEEE Journals}

\maketitle
\begin{abstract} The Metaverse, a burgeoning collective virtual space merging augmented reality and persistent virtual worlds, necessitates advanced artificial intelligence (AI) and communication technologies to support immersive and interactive experiences. Federated learning (FL) has emerged as a promising technique for collaboratively training AI models while preserving data privacy. However, FL faces challenges such as high communication overhead and substantial computational demands, particularly for neural network (NN) models. To address these issues, we propose an integrated federated split learning and hyperdimensional computing (FSL-HDC) framework for emerging foundation models. This novel approach reduces communication costs, computation load, and privacy risks, making it particularly suitable for resource-constrained edge devices in the Metaverse, ensuring real-time responsive interactions. Additionally, we introduce an optimization algorithm that concurrently optimizes transmission power and bandwidth to minimize the maximum transmission time among all users to the server. The simulation results based on the MNIST dataset indicate that FSL-HDC achieves an accuracy rate of approximately 87.5\%, which is slightly lower than that of FL-HDC. However, FSL-HDC exhibits a significantly faster convergence speed, approximately 3.733x that of FSL-NN, and demonstrates robustness to non-IID data distributions. 
Moreover, our proposed optimization algorithm can reduce the maximum transmission time by up to 64\% compared with the baseline.
\end{abstract}

\begin{IEEEkeywords}
Federated split learning (FSL), hyperdimensional computing (HDC), resource allocation.
\end{IEEEkeywords}

\IEEEpeerreviewmaketitle

\section{Introduction}



As the dawn of sixth-generation (6G) technology approaches, the Metaverse, a collective virtual shared space created by the convergence of augmented reality (AR) and persistent virtual worlds, promises to revolutionize human interaction with digital environments. The potential applications of the Metaverse span across diverse domains such as social networking, education, entertainment, and commerce, offering immersive and interactive experiences that transcend current technological boundaries \cite{zawish2024ai}. This expansive vision relies heavily on advancements in communication networks, computing, and artificial intelligence (AI).






Federated learning (FL) has emerged as a transformative approach that enables multiple devices to collaboratively train machine learning (ML) models without exchanging raw data. This paradigm ensures data privacy and addresses the issue of data silos. However, FL encounters significant challenges, primarily related to high communication overhead and substantial computational demands. Additionally, FL systems are vulnerable to various attacks, such as model poisoning, data poisoning, and deep leakage from gradients (DLG), which can maliciously compromise the global model or steal local user data. These security concerns further complicate the deployment of FL in sensitive applications. To address these challenges, split learning (SL) has been proposed as an innovative strategy. SL involves partitioning the model into two segments: one part is processed on the client side, and the other on the server side. This division effectively reduces the computational load on client devices and minimizes the data transmitted to the server \cite{10353003}. While SL alleviates some of the issues associated with FL, it still has room for improvement in terms of privacy and computational resource utilization.

Building on the strengths of both FL and SL, federated split learning (FSL) emerges as a hybrid approach that combines the benefits of both methodologies. FSL allows multiple clients to collaborate on training a model while distributing the computational tasks between the client and server. This integrated approach not only mitigates communication costs and lessens the computational demands on client devices but also significantly enhances the system's privacy security. By decentralizing data processing across multiple clients, FSL reduces the risks associated with centralized data storage and processing, thereby enhancing user data privacy protection.

Despite these advancements, the inherent drawbacks of utilizing neural networks (NNs) in FSL, such as their substantial computational resource demands and energy consumption, limit its applicability, especially in edge computing scenarios \cite{ma2024hyperdimensional}. To circumvent these challenges, hyperdimensional computing (HDC) \cite{kanerva2009hyperdimensional} presents itself as a compelling alternative. Inspired by the brain's way of processing information through high-dimensional representations, HDC offers significant advantages, including lower computational complexity, enhanced energy efficiency, and robustness. HDC encodes data into high-dimensional vectors that can be manipulated using simple arithmetic operations, making it highly suitable for deployment on edge devices with limited computational power. HDC's benefits of energy-efficient computation, ultra-fast one-pass training, and support for real-time learning and reasoning make it particularly suitable for Metaverse scenarios.

Considering the above-mentioned advantages of HDC, there has been a growing body of research on the use of HDC in distributed learning. For example, the work in \cite{zhang2023hyperdimensional} provided a systematic study on HDC-based FL under two HDC model aggregation strategies and investigated the impact of different parameter settings on the performance of FL-HDC. A novel HDC-based FL framework called HyperFeel was proposed in \cite{10473907}, which can significantly improve communication/storage
efficiency over existing works with nearly no performance degradation. Moreover,
an efficient FL-HDC framework was proposed in \cite{hsieh2021fl} by bipolarizing HVs to significantly reduce communication costs. 
The authors in \cite{morris2022} proposed an HDC system applied in the FL scenario named HyDREA, which adaptively changes the model bandwidth to maintain accuracy in high-noise transmission environments, achieving significant speedup and energy efficiency improvements.
However, the combination of FSL and HDC is still not investigated, even though FSL can further increase the privacy of HDC. 

In this paper, we introduce the integrated federated split learning and hyperdimensional computing (FSL-HDC) scheme, a novel framework designed to optimize the capabilities of FSL within the Metaverse. By integrating HDC into the FSL paradigm, we aim to tackle the computational and energy efficiency challenges faced by edge devices, facilitating more efficient and scalable learning across distributed environments. This innovative approach not only enhances the feasibility of deploying intelligent services in the Metaverse but also significantly improves the user experience through responsive and adaptive interactions. The low computational complexity of the FSL-HDC framework reduces the processing burden on edge devices, making it possible to implement complex learning algorithms in resource-constrained environments. By effectively distributing computational tasks between clients and servers, FSL-HDC can manage the massive data volumes generated in the Metaverse. The federated nature of FSL ensures that user data remains decentralized, reducing the risks associated with centralized data processing and significantly enhancing data privacy protection. The inherent robustness of HDC allows the system to adapt to dynamic changes in the environment, providing a seamless and intuitive user experience.

The contribution of the paper can be summarized as follows:
\begin{itemize}

\item To the best of our knowledge, we propose FSL-HDC as the first framework that integrates both FSL and HDC. FSL-HDC can partition the model into two parts based on the task's complexity and the user's conditions, such as energy, computing resources, and data diversity. One part is trained on the client-side using FL to obtain an intermediate HDC model, referred to as the smashed AM. Subsequently, further training is conducted on the server-side to obtain the final global HDC model.

\item To further improve the network performance, we propose an optimization problem that concurrently optimizes transmission power and bandwidth to minimize upload transmission time. Due to the fact that the transmission time between the fed server and the main server is negligible compared to the transmission time between the user and the fed server, in this work, we aim to minimize the maximum transmission time among all users to the fed server. To make the problem tractable, we propose an algorithm that solves transmission power optimization and bandwidth allocation in an alternating manner. Compared with the baseline, which only optimizes transmission power and uniformly allocates bandwidth, our joint optimization strategy can reduce transmission time by up to 64\%.
\item We evaluate the efficiency of FSL-HDC using the MNIST dataset under both IID and non-IID scenarios, and compare its performance against two baseline models. Simulation results demonstrate that the proposed FSL-HDC achieves an higher accuracy of approximately 87.5\%, compared to FL-HDC, while exhibiting a convergence speed that is 3.733x faster than FSL-NN. Additionally, FSL-HDC shows strong robustness against non-IID data, as training with non-IID data does not result in a significant loss of accuracy.
\end{itemize}




The organization of this paper is as follows. The HDC model is illustrated in Section \ref{HDC}. The proposed FSL-HDC model and problem formulation are given in Section \ref{model}. The algorithm design is given in Section \ref{AD}. Section \ref{simulation} describes the simulation results and analysis of the proposed approach, and the conclusion is summarized in Section \ref{con}.

\section{Hyperdimensional Computing  Model}\label{HDC}
In this section, we first introduce the basic concept of HDC, including its elements, operations, item memory, and associative memory. After that, the four stages of the HDC model, encoding, training, inference, and retraining, are presented. 

\subsection{Notions of HDC}
\subsubsection{Hypervector}
Hypervector (HV) is the fundamental element of HDC. It typically has dimensions exceeding 10000 with each dimension represented by a numerical value. These values are usually in binary, bipolar, or integer formats \cite{ge2020classification}. A $D$-dimensional HV can be expressed as ${H}=\left\langle h_1, h_2, \ldots, h_D\right\rangle$, where $h_{i}$ represents the value in the 
${i}$-th dimension. In this context, all dimensions are considered to contribute equally. Additionally, HVs are usually randomly initialized, with the generated values following independent and identically distributed (IID) randomness. Due to the high-dimensional nature of HVs, this ensures that two randomly initialized HVs are almost orthogonal to each other \cite{ma2024hyperdimensional}.

\subsubsection{Operations}
Operations on HVs are designed to manipulate and combine them while preserving and leveraging their high-dimensional characteristics. These operations facilitate the representation and processing of complex information within HDC systems. It is crucial to note that these operations do not alter the dimensions of the HV operands. The primary operations include:

\begin{itemize}
\item Addition (Bundling): This operation combines multiple HVs into a single HV, representing the collective information of the inputs. Addition is typically performed element-wise. 
\item Multiplication (Binding): Binding is used to create associations between HVs. It typically involves element-wise multiplication (e.g., XOR for binary HVs or element-wise multiplication for real-valued and bipolar HVs). This operation produces a new HV that represents the relationship between the input vectors.
\item Permutation $\rho$: Permutation is an operation that reorders HV, generating a new HV $\rho(H)$ by altering the sequence of elements in the original HV. The resultant vector is quasi-orthogonal to the original, implying a normalized hamming distance close to 0.5, where approximately half of the bits differ. This quasi-orthogonality ensures that the permuted $\rho({H})$ is statistically distinct from the original $H$ while maintaining a degree of relatedness. This distinction enhances computational accuracy and robustness. Permutation can be implemented through a permutation matrix, with circular shift as a commonly employed hardware-friendly method \cite{ge2020classification}.


\end{itemize}


Due to the inherent characteristics of these three operations, HDC results in rapid learning capabilities, high energy efficiency, and acceptable accuracy in both learning and classification tasks \cite{ge2020classification}.

\subsubsection{Item Memory}
Item Memory (IM) in HDC is a repository that stores predefined HVs corresponding to specific items or symbols. Each item in the IM is associated with a unique HV, which is usually randomly generated and then fixed. These HVs serve as the building blocks for representing more complex data structures. The main purpose of the IM is to provide a consistent and efficient way to encode basic elements, ensuring that each element has a unique, high-dimensional representation that can be easily retrieved and used in further computations.
\subsubsection{Assocoative Memory}
Associative Memory (AM) in HDC is designed to store and retrieve patterns based on similarity. Unlike traditional memory systems that rely on exact matches, AM uses similarity measures to find the closest match to a given query HV. This capability is particularly useful for tasks involving pattern recognition, classification, and noisy data retrieval. During training, class-specific prototype HVs are stored in the AM. During inference, the encoded input HV is compared against these prototypes, and the closest match is retrieved, effectively classifying the input.
\subsection{HDC Model Development}

\subsubsection{Encoding}
\begin{figure*}[t]
 \centering\includegraphics[width=0.95\textwidth]{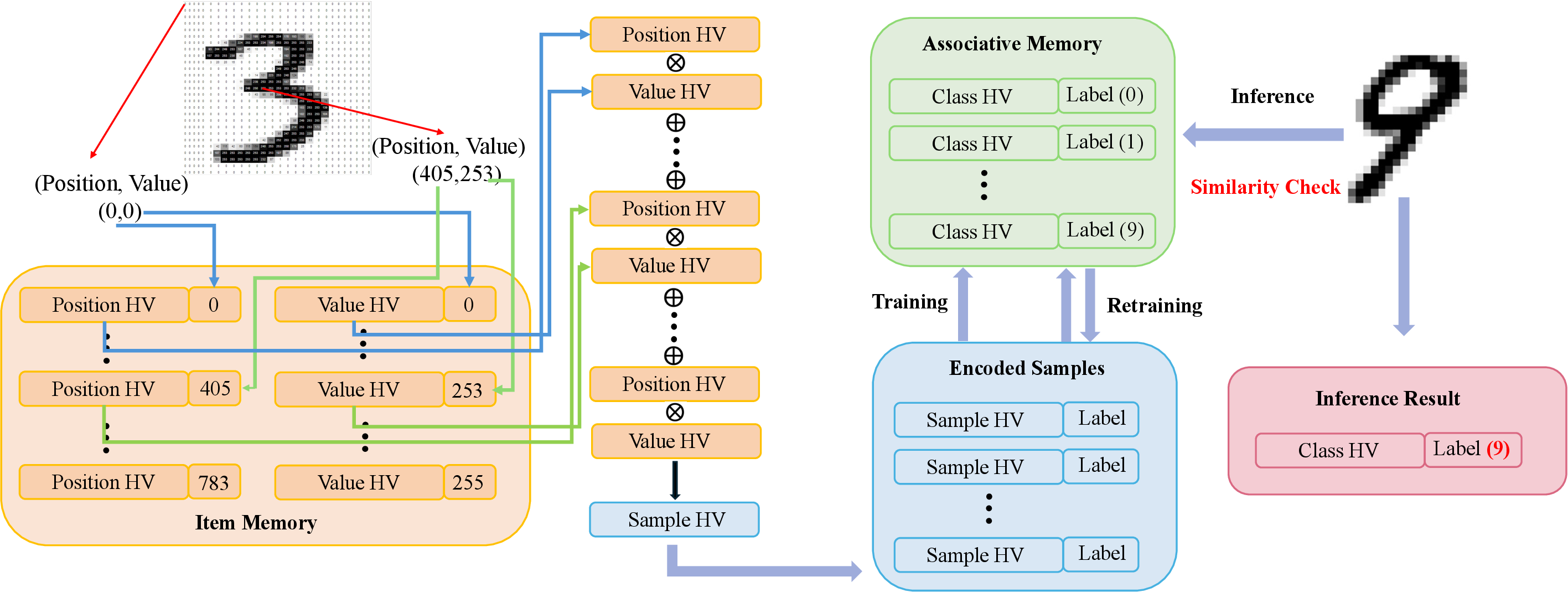}
    \caption{Overview of HDC model encoding, training, inference, and retraining.}
    \label{fig:H1}
\end{figure*}

The encoding process in HDC resembles feature extraction from input samples, embedding them into HVs. Two primary encoding methods are commonly used: record-based encoding \cite{rahimi2016hyperdimensional} and n-gram-based encoding \cite{joshi2017language}. We elucidate the detailed processes using the record-based encoding method, as illustrated in Fig. \ref{fig:H1}. 

Firstly, two types of IMs are generated: one is iM, which stores the feature position HVs, termed as $H_p$; the other is continuous item memory (CiM), which stores the feature value HVs, term as $H_v$. Using the MNIST image dataset as an example, each image is divided into 784 pixels. During the encoding process, 784 position HVs are randomly generated to encode the feature positions, and 256 value HVs are generated to represent pixel values (ranging from 0 to 255). For each pixel, its position HV and value HV are multiplied, and by adding all these multiplied results from each pixel in the sample image, an HV representing this sample, termed the sample HV or query HV, is generated  as
\begin{equation}
\begin{aligned}
H_m=\sum_{k=0}^{783} H_{m,p(k)} \odot H_{m,v(pixel(k))},
\label{encoding}
\end{aligned}
\end{equation}
where $m$ refers to the $m$-th sample, $H_{m,p(k)}$ denotes the HV at position $k$ in the $m$-th sample, and $H_{m,v(pixel(k))}$ represents the value of the pixel corresponding to position $k$ in the HV. $\odot$ is the element-wise product.



\subsubsection{Training}
Training involves adding all encoded sample HVs of the same class $n$ to obtain the class HVs $A^n$, and then aggregating all the class HVs to form the AM $\mathcal{A}$.
Mathematically, $A^n$ and $\mathcal{A}$ can be respectively calculated as 
\vspace{-0.5em}
\begin{equation}
A^n=\sum_{j=1}^{k_n} H_{n_j},
\label{A1}
\end{equation}
\vspace{-1em}
\begin{equation}
\mathcal{A}=\left\{{A^1}, {A^2}, \ldots, {A^N}\right\},
\label{A2}
\end{equation}
where $k_n$ denotes the number of encoded training data in the class $n$, $ H_{n_j}$ denotes the $j$-th HV in the class $n$, and $N$ represents the total number of classes. For example, in the MNIST dataset, $N = 10$.

It is important to note that the basic training process of HDC does not require multiple iterations like NN models. One significant feature of HDC is that it can achieve reasonable accuracy with just a single training pass.


\subsubsection{Inference}
Inference in HDC is the process by which the trained model classifies input data. When an input is presented, it is first encoded into a query HV $H_q$ using the same encoding process as during training. This query HV is then compared to each class HVs in AM using a similarity measure, such as cosine similarity given by 
\begin{equation}
\begin{array}{r}
\xi_{\text {cosine }}\left({H}_q, {A}^i\right)=\frac{{H}_q \cdot {A}^n}{\left\|{H}_q\right\| \times\left\|{A}^n\right\|} \;,
\label{cos}
\end{array}
\end{equation}
where $\cdot$ denotes the dot product, and $\left\|{H}_q\right\|$ and $\left\|{A}^n\right\|$ present the Euclidean norm of ${H}_q$ and the HV of class $n$, respectively.

The class with the highest similarity score is assigned as the predicted class for the input. This method ensures that the classification decision leverages the high-dimensional properties of the HVs, providing robustness against noise and variability in the input data.

\subsubsection{Retraining}

To further improve the accuracy of the HDC model, retraining can be applied to the AM over several additional iterations on the training set. During retraining, each sample HV is classified. If the prediction is correct, no adjustment is necessary. If the predicted label does not match the true label, the corresponding query HV $H_q$
is subtracted from the misclassified class $A^{\text{miss}}$
and simultaneously added to the real class $A^{\text {real}}$. The process is as follows:
\begin{equation}
\begin{array}{r}
{A^{{\text {miss}}}}={A^{H^{\text {miss}}}}-\alpha {H_q}\;, \\
{A^{\text {real}}}={A^{\text {real}}}+\alpha{H_q}\;,
\end{array}
\label{retraining}
\end{equation}
where $\alpha$ is the learning rate.

\section{HDC-Based FSL model and Problem Formulation}\label{model}
In this section, we introduce the proposed FSL-HDC system model, transmission model, and optimization problem.

\subsection{System Model}
In our FSL-HDC system, we have one BS as the main server, one edge server as the federated server (fed server), and $U$ clients, where the fed server is close to the clients. 
\subsection{The Framework of HDC based FSL}
\begin{figure*}[t]
 \centering\includegraphics[width=0.85\textwidth]{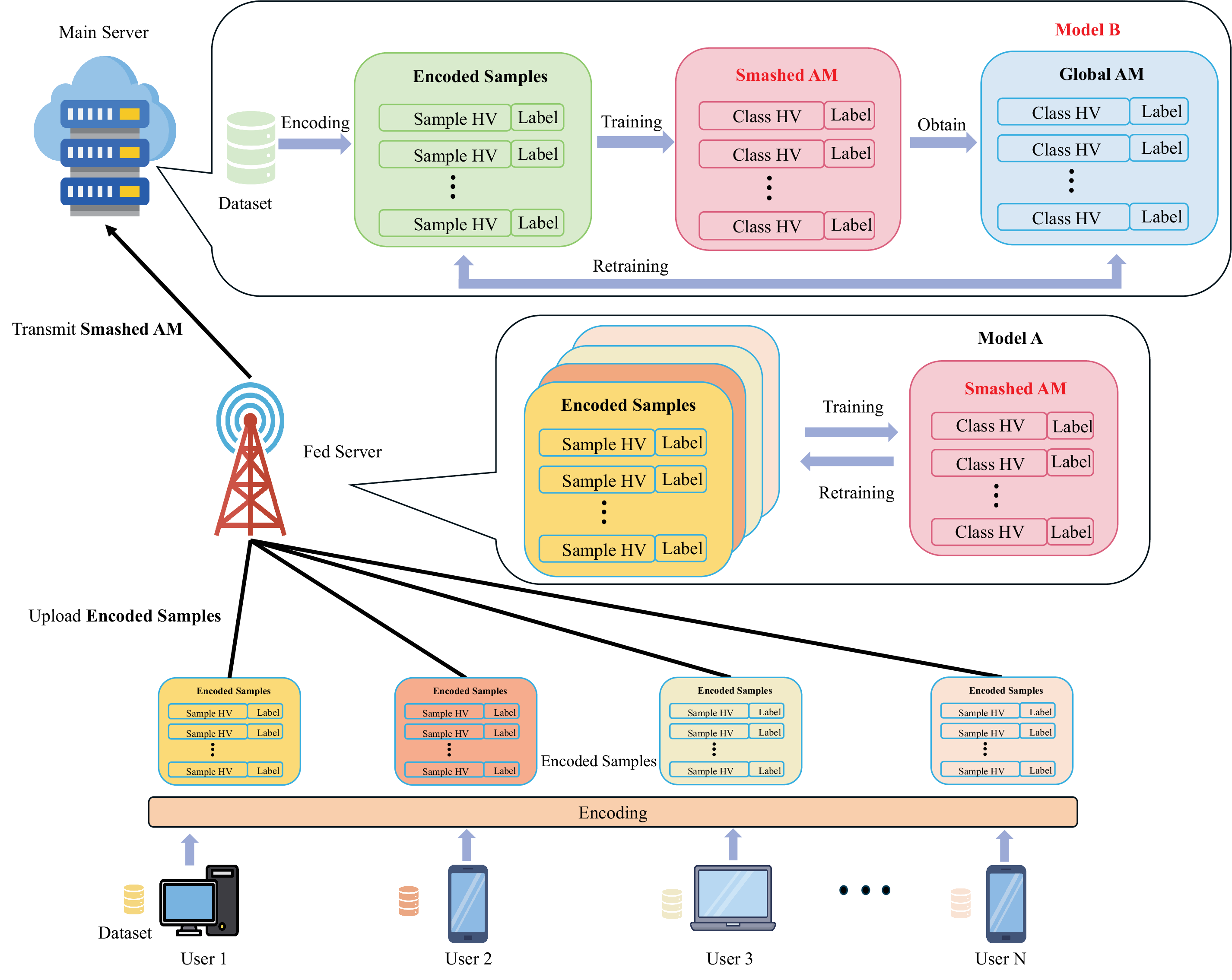}
    \caption{The architecture of proposed FSL-HDC framework.} 
    \label{fig:H2}
\end{figure*}

The FSL-HDC framework leverages the key advantages of FL, SL, and HDC. It achieves parallel processing of distributed clients by partitioning the model into client-side and server-side submodels during training. Additionally, it capitalizes on the simplicity, compactness, and efficiency of HDC operations, making it suitable for edge devices. As illustrated in Fig. \ref{fig:H2}, this approach ensures efficient training and inference in resource-constrained environments. 

In SL framework, to collaborately train the whole model between the two ends, the entire model is divided into model $A$ and model $B$ based on the task and clients' dataset. Clients and the fed server collaboratively train model $A$, and the AM of model $A$ is transmitted as smashed data to the main server. The main server then uses these smashed data and its own dataset to train model $B$, ultimately producing the final global model AM, which is broadcast to all clients. Note that we use D-dimensional HVs with bipolar elements $\{+1,-1\}^D$. The details of the FSL-HDC training process include three steps and the whole process can be described in Algorithm~ \ref{ag1}:

\begin{algorithm}
\caption{HDC-based SFL} 
\begin{algorithmic}[1]
\State Client local samples $\mathcal S$; Main server local samples $\mathcal K$; Retraining epoch in fed server $E$; Retraining epochs in main server $N$.
\State \textbf{Client executes:}
\For {each client do}
    \State $\text{sample} HV_{\text{client}} \gets \text{client.encode}(\mathcal{S})$
    \State \text{client.upload} ($\text{sample} HV_{\text{client}}$)
\EndFor
\State

\State \textbf{Fed server executes:} \textcolor{blue}{\Comment{training Model A at fed server side}}
\State $\mathcal{A}_{\text{smashed}} \gets \sum \text{sample} HV_{\text{client}}$
\State $\mathcal{A}_{\text{smashed}} \gets \text{fed server.retrain}(\mathcal{A}_{\text{smashed}}, E)$
\State $\text{fed server.upload}  
 (A_{\text{smashed}})$
\State
\State \textbf{Main server executes:}\textcolor{blue}{\Comment{training Model B at main server side}}
\State $\text{sample} HV_{\text{main server}} \gets \text{main server.encode}(\mathcal{K})$
\State $\mathcal{A}_{\text{global}} \gets \text{main server.train}(\mathcal{A}_{\text{smashed}}, \text{sample} HV_{\text{main server}})$
\State $\mathcal{A}_{\text{global}} \gets \text{main server.retrain}(\mathcal{A}_{\text{global}},E,\alpha)$
\For {each client do}
    \State $\mathcal{A}_{\text{global}} \gets \text{client.download}(\mathcal{A}_{\text{global}})$
\EndFor
\end{algorithmic}
\label{ag1}
\end{algorithm}
\textbf{(1) Encoding on the Clients:} The main server first generates bipolar HV benchmarks to ensure consistency in the position HVs and value HVs across all clients. Each client then encodes their local data into sample HVs using Eq.\eqref{encoding}, and uploads all sample HVs to the fed server for training model $A$.




\textbf{(2) Submodel training on fed server:} 
The fed server aggregates the sample HVs from clients by class to obtain class HVs $A^n$ and the $\mathcal{A}$. The HDC model can then undergo retraining based on the accuracy achieved during the one-pass training. During retraining, all query HVs are compared with each class $A^n$ in the $\mathcal{A}$ using similarity measure, as shown in Eq.\eqref{cos}. For incorrectly predicted cases, the class HVs are updated according to Eq.\eqref{retraining}. After the fed server completes retraining, each retrained class HV is bipolarized to obtain the smashed AM $\mathcal{A}_{smashed}$, which is then uploaded to the main server.

\textbf{(3) Further training on the main server:} 
The main server first encodes its own local $K$ datasets into sample HVs. Using these local sample HVs along with the received $\mathcal{A}_{smashed}$, the main server trains model $B$. The main server then performs retraining on the global AM by iterating through its local sample HVs until the model converges. Once the final global AM is obtained, it is broadcast to all clients. Clients will subsequently use this distributed AM for their inferences.

\subsection{Transmission Model}
As shown in Fig. \ref{fig:H2}, a fed server-assisted network is considered in this work, where the fed server with edge computing capability is assumed to be geographically located at the network center. Users are randomly distributed within this network, and the set of all users is denoted as $U$. As mentioned earlier, the transmission time between the fed server and the main server is negligible compared to the transmission time between the user and the fed server. Therefore, in this work, we aim to minimize the maximum transmission time among all users to the fed server. 
To estimate transmission properties, we first denote the line of sight (LoS) probability for a mmWave link between a user and the fed server with distance $d$ as ${\mathbb{P}}_{\text{LoS}}(d)$. We employ the model for urban macro scenarios in \cite{rappaport2017overview} as follows:

\begin{equation}
    {\mathbb{P}}_{\text{LoS}}(d)=\left \{
    \begin{array}{llc}
        1 &  d\le 18({\text{m}}) \\
        \frac{18}{d}+{\text{exp}}(-\frac{(d-18)}{63})(1-\frac{18}{d}) & d>18(\text{m}) 
    \end{array}
       \right.,
\end{equation}
where the non-line of sight (NLoS) probability is $(1-{\mathbb{P}}_{\text{LoS}}(d))$, and the path loss model is given as follows:
\begin{equation}
\begin{aligned}
    L(d) &= {\mathbb{P}}_{\text{LoS}}(d)\cdot L_{\text{LoS}}(d)+(1-{\mathbb{P}}_{\text{LoS}}(d))\cdot L_{\text{NLoS}}(d)\\
    &={\mathbb{P}}_{\text{LoS}}(d)\cdot \beta d^l+(1-{\mathbb{P}}_{\text{LoS}}(d))\cdot\beta d^n,
    \end{aligned}
\end{equation}
where $L_{\text{LoS}}(d)$ and $L_{\text{NLoS}}(d)$ refer to path loss for LoS and NLoS link, respectively. $\beta=\frac{4\pi dc}{f_c}$ refers to the reference-free space path loss at 1 m distance. $l$ and $n$ refer to path loss exponents for LoS and NLoS links, respectively.

Next, the achievable transmission rate between a user $i$ and the fed server with the allocated bandwidth is given as follows:
\begin{equation}
    r_i=b_i\log_2\left(1+\frac{p_i}{n_0b_i\cdot L_i(d)}\right),
\end{equation}
where $n_0b_i$ refers to the white Gaussian noise, and $b_i$ refers to the bandwidth allocated to user $i$. Additionally, $p_i$ refers to the transmission power. Hereafter, the transmission time between the user and the fed server, to transmit a file with a size of $Q_i$ bits is as follows:
\begin{equation}
    t_i = \frac{Q_i}{r_i},\quad\forall i\in U.\label{time_i}
\end{equation}

In this work, the fed server conducts training after receiving sample HVs from all users. Therefore, to further improve network performance, we aim to minimize the maximum transmission delay between users and the fed server while satisfying the energy constraints of all users. To simplify the optimization process, we consider this is a synchronous system, in other words, the users are  to start transmitting sample HVs at the same time \cite{yang2020energy}, and the optimization problem is formulated as follows:
\begin{subequations}
\begin{align}
\text{P1:}\qquad &\min_{_{p_i,b_i}}\max_{i} \quad t_i\label{obj}\\
\text{s.t.}
\quad& p_i\leq P_{\text{max}}, \;\forall i\in U,\label{pmax}\\
\;& \sum_{\forall i\in U}b_i\leq B,\label{Nmax}\\
\;& t_i\cdot p_i\leq E, \;\forall i\in U,\label{Emax}\\
\;& p_i\ge 0, \;\forall i\in U,\label{ppositive}\\
\;& b_i\ge 0\label{binteger},
\end{align}
\end{subequations}
where the problem objective in \eqref{obj} aims to minimize the maximum transmission time among all users. The constraint \eqref{pmax} limits the maximum transmission power of each user. The constraint \eqref{Nmax} refers to that the sum of all allocated bandwidth is smaller than the total amount of bandwidth. Next, the constraint \eqref{Emax} is the transmission energy limitation. Lastly, constraints \eqref{ppositive} and \eqref{binteger} define the type of variables.

\section{Algorithm Design}\label{AD}
To make the transmission problem P1 tractable, first, we replace the $Q_i$ of each user with $Q$ in \eqref{time_i}. This is due to the fact that after encoding procedures, the transmitted file, that is, the HV, is characterized by the same size for each user. Next, we introduce an auxiliary variable $T$, and transform the original problem into the problem shown in P2 without loss of optimality.
\begin{subequations}
\begin{align}
\text{P2:}\qquad &\min_{_{p_i,b_i,T}} \quad T\\
\text{s.t.}
\quad& \eqref{pmax}-\eqref{binteger}\nonumber\\
\;& T\geq \frac{Q}{r_i},\;\forall i\in U,\\
\;& T\geq 0.
\end{align}
\end{subequations}

Next, we replace variable $T$ with auxiliary variable $\gamma$, where $\gamma=\frac{1}{T}$. Problem P2 is transformed into the following problem,

\begin{subequations}
\begin{align}
\text{P3:}\qquad &\max_{_{p_i,b_i,\gamma}} \quad \gamma\\
\text{s.t.}
\quad& \eqref{pmax}-\eqref{binteger}\nonumber\\
\;& Q\gamma\leq {r_i},\;\forall i\in U.\\
\;& \gamma\geq 0.
\end{align}
\end{subequations}

It can be observed that problem P3 is a non-convex problem because of the non-convex constraints. To obtain a suboptimal solution, we propose an algorithm that solves power optimization and bandwidth allocation in an alternating manner.
\subsubsection{Transmission power optimization}
With a fixed spectrum allocation solution, problem P3 can be reformulated as follows:
\begin{subequations}
\begin{align}
\text{P3-1:}\qquad &\max_{_{p_i,\gamma}} \quad \gamma\\
\text{s.t.}
\quad& p_i\leq P_{\text{max}}, \;\forall i\in U,\\
\;& \frac{b_i\log_2(1+\frac{p_i}{n_0b_i\cdot L_i(d)})}{p_i}\geq \frac{Q}{E}, \;\forall i\in U,\label{functionp}\\
\;& {b_i\log_2(1+\frac{p_i}{n_0b_i\cdot L_i(d)})}\geq Q\gamma,\;\forall i\in U,\\
\;& \gamma\geq 0,\\
\;&p_i\geq 0,\;\forall i\in U.
\end{align}
\end{subequations}

It is clear that the value of $\gamma$ in P3-1 is monotonically increasing with the value of transmission power. To find the optimal objective value, the maximum optimal transmission power within the feasible region is identified in the following proposition.

\textit{Proposition 1}: Given the bandwidth allocation, the optimal transmission power of each user is given by
\begin{equation}
p_i^{\star}=\min\{P_{\text{max}}, \hat{p}_i\},
\end{equation}
where $\hat{p}_i$ is the value obtained from the following equation: $\frac{b_i\log_2(1+\frac{\hat{p}_i}{n_0b_i\cdot L_i(d)})}{\hat{p}_i}= \frac{Q}{E}$.

\textit{Proof}: Since the objective function is an increasing function w.r.t. $p_i$, the optimal $p_i^{\star}$ is the maximum value within the feasible region. Let the left-hand-side of constraint \eqref{functionp} is denoted as $f(p_i)$, $f(p_i)=\frac{b_i\log_2(1+\frac{p_i
}{n_0b_i\cdot L_i(d)})}{p_i}$. It can be observed that $f(p_i)$ is a decreasing function w.r.t. $p_i$. Hence, the maximum feasible transmission power defined in constraint \eqref{functionp} satisfies condition $\frac{b_i\log_2(1+\frac{\hat{p}_i}{n_0b_i\cdot L_i(d)})}{\hat{p}_i}= \frac{Q}{E}$. This completes the proof.
\subsubsection{Bandwidth optimization}
With a fixed user transmission power solution, problem P3 can be reformulated as follows:
\begin{subequations}
\begin{align}
\text{P3-2:}\qquad &\max_{_{b_i,\gamma}} \quad \gamma\\
\text{s.t.}
\quad& \frac{b_i\log_2(1+\frac{p_i}{n_0b_i\cdot L_i(d)})}{p_i}\geq \frac{Q}{E}, \;\forall i\in U,\\
\;& {b_i\log_2(1+\frac{p_i}{n_0b_i\cdot L_i(d)})}\geq Q\gamma,\;\forall i\in U,\\
\;& \sum_{\forall i\in U}b_i\leq N,\\
\;& \gamma\geq 0,\\
\;& b_i\geq 0.\label{RBsubproblem}
\end{align}
\end{subequations}

It is clear problem P3-2 is a convex problem and can be solved optimally by using the standard convex optimization tool.
\section{Simulation Results and Analysis}\label{simulation}

\subsection{Experimental Setups for FSL-HDC}
In our experiments, we utilize MNIST \cite{lecun1998gradient} to demonstrate the effectiveness of our proposed FSL-HDC framework. We evaluate the testing accuracy of the global HDC model by implementing an image digit odd-even classification task. This setup involves 10 clients who interact with a fed server to train a digit classification model. The intermediate AM is then transmitted to the main server for further training to achieve the final odd-even classification.
We randomly select 6000 images (600 images per class) from the MNIST training set as the clients' training data, with each client's dataset consisting of 600 images. Additionally, we randomly select 2000 images as the main server's dataset. For the clients, we consider both IID and non-IID data distributions. For IID, 60 images from each class are randomly assigned to each user without overlap. For non-IID, we follow the data partition method described in \cite{mcmahan2017communication}: all data are first sorted, then divided into 20 shards, and each client is randomly assigned two shards. Moreover, we set the dimension of HV $D = 10000$, the number of epochs for retraining on the fed server is 15, and the learning rate is set to $\alpha=1$.

We compare the performance of FSL-HDC with two baseline models:
\begin{itemize}
\item FL-HDC: This baseline method employs record-based encoding and HV aggregation similar to our proposed method. Clients only encode their dataset into sample HVs, which are then trained and retrained by the fed server to produce the final global AM.
\item FSL-NN: The structure of this baseline closely mirrors our proposed method. Initially, clients and the fed server use the FedAvg algorithm to train a digit classification model over 100 epochs. Subsequently, the smashed data (intermediate layer) is transmitted to the main server for further training of the parity classification model.
\end{itemize}

\subsection{Experimental Setups for Optimization Problem}

To further improve the transmission process between users and the fed server, the maximum transmission time among all users is minimized via the joint optimization of transmission power and spectrum bandwidth. Simulations are conducted within a 200 m $\times$ 200 m area. The key notations and parameter values employed in this work are summarized in Table \ref{Notation}. We compare our proposed joint optimization with the baseline under different $P_{max}$ constraints. The baseline refers to optimizing only the users' transmission power while uniformly allocating bandwidth to all users.

\begin{table}[t]
\caption{Simulation parameters}
\begin{center}
    \begin{tabular}{lc}
    \hline
    \textbf{Parameter} &\textbf{value}\\
    \hline
    \hline
    $f_c$: carrier frequency    & 28 GHz \\
    $B$: total spectrum bandwidth & 100-500 MHz \\
    $P^{max}$: maximum transmission power  &  1 w\\
    $E$: maximum transmit energy & 5 J\\
    $n_0$: white Gaussian noise power spectral density  & -174 dBm/Hz\\
    $l$,$n$: path loss index for LoS and NLoS links & 2, 3.3 \\
     \hline
    \end{tabular}
    \label{Notation}
    \end{center}
\end{table}


\begin{figure}[t]
\centering\includegraphics[width=0.48\textwidth]{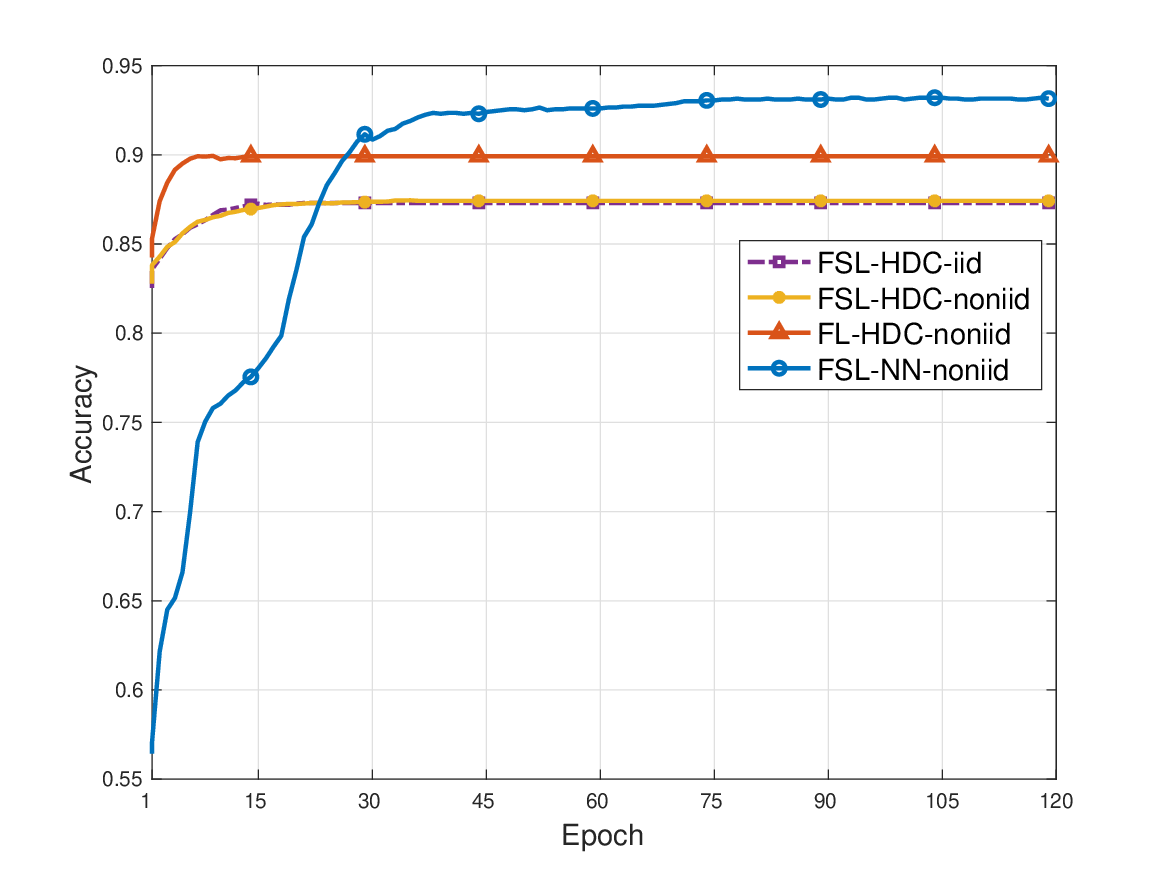}
    \caption{Accuracy vs. Epochs with different data distribution of FSL-HDC and non-IID of two baselines.}
    \label{fig:111}
\end{figure}

\subsection{Effects of Different Data Distribution}

In NN models, uneven data distribution can significantly impact training accuracy. To investigate the effect of data distribution on the FSL-HDC model, we evaluated the final model accuracy under identical conditions with user data being either IID or non-IID. As illustrated in Fig. \ref{fig:111}, the data distribution has a negligible effect on the FSL-HDC model, achieving approximately 87.5\% accuracy in both scenarios. Furthermore, the two models converge after approximately 10 retraining rounds. This stability is attributed to the HV aggregation method employed in FSL-HDC, where users encode their local data and upload them to the fed server. The fed server aggregates all encoded sample HVs during training. Consequently, the distribution of data among users does not affect the trained AM on the fed server.

\subsection{Comparision with Baselines}

Fig. \ref{fig:111} depicts the convergence rates and accuracies of three distinct algorithms. We compared our proposed FSL-HDC model with the FSL-NN and FL-HDC models under non-iid conditions.
Compared to FSL-NN, the FSL-HDC method demonstrates a convergence speed of approximately 3.733x faster. This advantage is due to the inherent differences in the training processes of HDC and NN models. HDC models typically exhibit lower complexity, utilizing simple linear or nonlinear transformations. Unlike NN models, the HDC models do not require intricate gradient computations and backpropagation during training, resulting in reduced computational overhead and faster training speeds. In most cases, a one-pass training suffices to achieve reasonable accuracy. Although the NN model achieves slightly higher accuracy after convergence, our algorithm eliminates the need for local iterative training and user-server interaction, making it particularly suitable for users with limited computational resources who are unable to train a global model.

When compared to FL-HDC, the convergence speed of the FSL-HDC model is comparable, with a minor accuracy reduction of 2.33\%. However, our method offers increased security, rendering it more suitable for complex inference tasks, especially when local user data is insufficient and certain data is exclusively available on the main server.

\subsection{Transmission time minimization}

\begin{figure}[t]
\centering\includegraphics[width=0.48\textwidth]{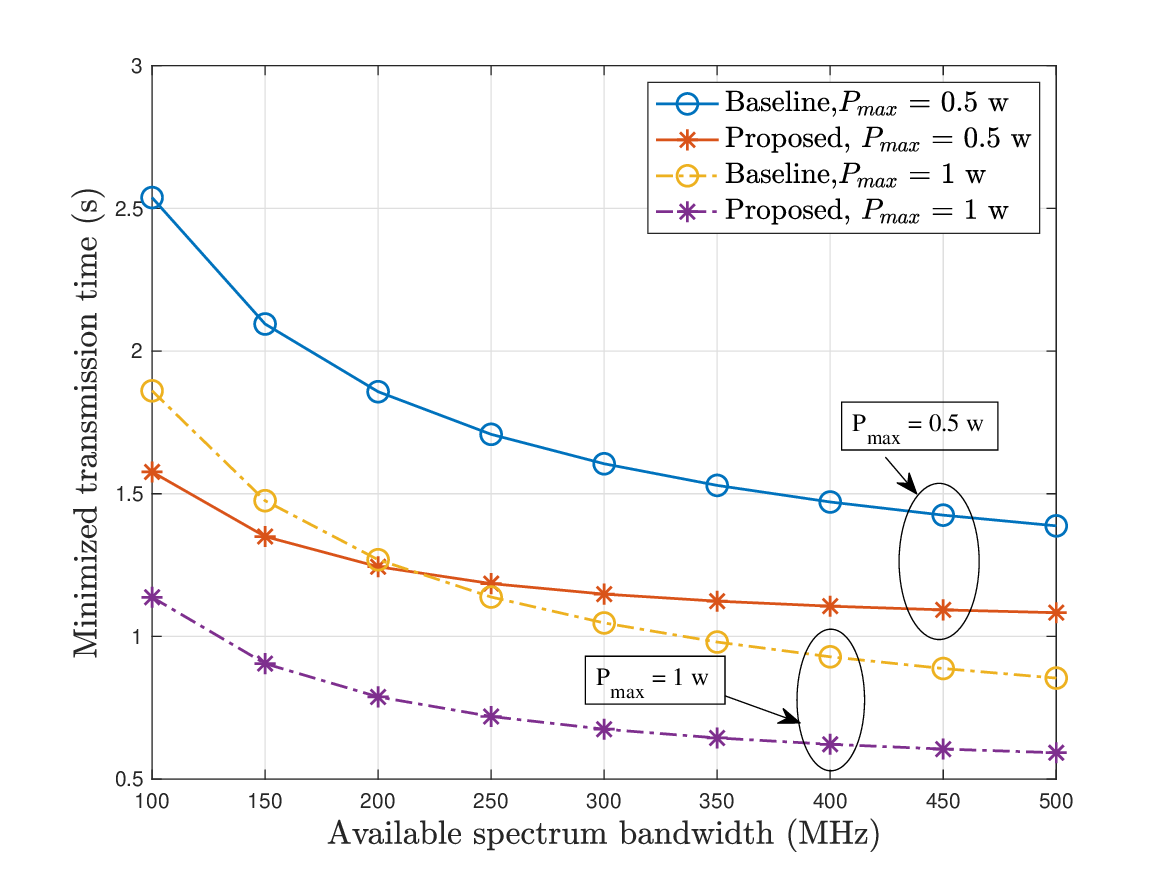}
    \caption{Minimized transmission time under different available spectrum bandwidth resources.}
    \label{fig:mint}
\end{figure}

Fig. \ref{fig:mint} compares the optimized results of minimized transmission time of the proposed algorithm with baseline results. The baseline model optimizes transmission power while the bandwidth is evenly distributed among all users. It can be observed that the transmission time decreases with the increase of available bandwidth resources and maximum transmission power. Additionally, the proposed algorithm outperforms the baseline results. For instance, the proposed algorithm can reduce transmission time by up to $60\%$ and $64\%$ compared to the baseline results when the maximum transmission power is 0.5 w and 1 w, respectively. Moreover, it can be observed that, compared to the baseline result, the performance improvement is more pronounced when the bandwidth resource is less abundant.

\section{Conclution}\label{con}
In this paper, we proposed a novel FSL-HDC scheme designed for resource-limited devices in the Metaverse. This framework integrates the advantages of FSL and HDC, effectively reducing communication costs and computation overhead while enhancing efficiency, privacy, and robustness against non-IID data distributions. Additionally, we introduced an optimization algorithm aimed at minimizing the maximum transmission delay from users to the fed server. Experimental results demonstrated that our proposed FSL-HDC scheme exhibits strong robustness to non-IID data and significantly outperforms traditional neural network model FSL-NN in terms of convergence speed. The accuracy is slightly lower than that of FL-HDC. Moreover, the proposed optimization method substantially reduces the maximum transmission time, thereby further accelerating the model's convergence rate. For instance, compared with the baseline, our joint optimization of transmission power and bandwidth can reduce transmission time by up to 64\%.



%


\ifCLASSOPTIONcaptionsoff
  \newpage
\fi



%

%



\bibliographystyle{IEEEtran}
\bibliography{IEEEabrv,ref}

\begin{thebibliography}{10}
\providecommand{\url}[1]{#1}
\csname url@samestyle\endcsname
\providecommand{\newblock}{\relax}
\providecommand{\bibinfo}[2]{#2}
\providecommand{\BIBentrySTDinterwordspacing}{\spaceskip=0pt\relax}
\providecommand{\BIBentryALTinterwordstretchfactor}{4}
\providecommand{\BIBentryALTinterwordspacing}{\spaceskip=\fontdimen2\font plus
\BIBentryALTinterwordstretchfactor\fontdimen3\font minus \fontdimen4\font\relax}
\providecommand{\BIBforeignlanguage}[2]{{%
\expandafter\ifx\csname l@#1\endcsname\relax
\typeout{** WARNING: IEEEtran.bst: No hyphenation pattern has been}%
\typeout{** loaded for the language `#1'. Using the pattern for}%
\typeout{** the default language instead.}%
\else
\language=\csname l@#1\endcsname
\fi
#2}}
\providecommand{\BIBdecl}{\relax}
\BIBdecl

\bibitem{zawish2024ai}
M.~Zawish, F.~A. Dharejo, S.~A. Khowaja, S.~Raza, S.~Davy, K.~Dev, and P.~Bellavista, ``{AI and 6G into the metaverse: Fundamentals, challenges and future research trends},'' \emph{IEEE Open J. Commun. Soc.}, vol.~5, pp. 730--778, 2024.

\bibitem{10353003}
Y.~Ding, Z.~Yang, Q.-V. Pham, Y.~Hu, Z.~Zhang, and M.~Shikh-Bahaei, ``{Distributed machine learning for UAV Swarms: Computing, sensing, and semantics},'' \emph{IEEE Internet Things J.}, vol.~11, no.~5, pp. 7447--7473, 2024.

\bibitem{ma2024hyperdimensional}
D.~Ma, C.~Hao, and X.~Jiao, ``{Hyperdimensional computing vs. neural networks: Comparing architecture and learning process},'' in \emph{Proc. Int. Symp. Qual. Electron. Des. (ISQED)}.\hskip 1em plus 0.5em minus 0.4em\relax IEEE, 2024, pp. 1--5.

\bibitem{kanerva2009hyperdimensional}
P.~Kanerva, ``{Hyperdimensional computing: An introduction to computing in distributed representation with high-dimensional random vectors},'' \emph{Cognit. Comput.}, vol.~1, pp. 139--159, 2009.

\bibitem{zhang2023hyperdimensional}
S.~Zhang, D.~Ma, S.~Bian, L.~Yang, and X.~Jiao, ``{On hyperdimensional computing-based federated learning: A case study},'' in \emph{Proc. Int. Joint Conf. Neural Netw.(IJCNN)}.\hskip 1em plus 0.5em minus 0.4em\relax IEEE, 2023, pp. 1--8.

\bibitem{10473907}
H.~Li, F.~Liu, Y.~Chen, and L.~Jiang, ``{HyperFeel: An efficient federated learning framework using hyperdimensional computing},'' in \emph{Proc. Asia South Pacific Des. Autom. Conf. (ASP-DAC)}, 2024, pp. 716--721.

\bibitem{hsieh2021fl}
C.-Y. Hsieh, Y.-C. Chuang, and A.-Y.~A. Wu, ``{Fl-hdc: Hyperdimensional computing design for the application of federated learning},'' in \emph{Proc. IEEE Int. Conf. Artif. Intell. Circuits Syst. (AICAS)}.\hskip 1em plus 0.5em minus 0.4em\relax IEEE, 2021, pp. 1--5.

\bibitem{morris2022}
J.~Morris, K.~Ergun, B.~Khaleghi, M.~Imani, B.~Aksanli, and T.~Simunic, ``{Hydrea: Utilizing hyperdimensional computing for a more robust and efficient machine learning system},'' \emph{ACM Trans. Embed. Comput. Syst.}, vol.~21, no.~6, pp. 1--25, 2022.

\bibitem{ge2020classification}
L.~Ge and K.~K. Parhi, ``{Classification using hyperdimensional computing: A review},'' \emph{IEEE Circuits Sys. Mag.}, vol.~20, no.~2, pp. 30--47, 2020.

\bibitem{rahimi2016hyperdimensional}
A.~Rahimi, S.~Benatti, P.~Kanerva, L.~Benini, and J.~M. Rabaey, ``{Hyperdimensional biosignal processing: A case study for EMG-based hand gesture recognition},'' in \emph{Proc. IEEE Int. Conf. Rebooting Comput. (ICRC)}.\hskip 1em plus 0.5em minus 0.4em\relax IEEE, 2016, pp. 1--8.

\bibitem{joshi2017language}
A.~Joshi, J.~T. Halseth, and P.~Kanerva, ``{Language geometry using random indexing},'' in \emph{Proc. Int. Symp. Quantum Interaction}.\hskip 1em plus 0.5em minus 0.4em\relax Springer, 2017, pp. 265--274.

\bibitem{rappaport2017overview}
T.~S. Rappaport, Y.~Xing, G.~R. MacCartney, A.~F. Molisch, E.~Mellios, and J.~Zhang, ``{Overview of millimeter wave communications for fifth-generation (5G) wireless networks—With a focus on propagation models},'' \emph{IEEE Trans. Antennas Propag.}, vol.~65, no.~12, pp. 6213--6230, 2017.

\bibitem{yang2020energy}
Z.~Yang, M.~Chen, W.~Saad, C.~S. Hong, and M.~Shikh-Bahaei, ``Energy efficient federated learning over wireless communication networks,'' \emph{IEEE Transactions on Wireless Communications}, vol.~20, no.~3, pp. 1935--1949, 2020.

\bibitem{lecun1998gradient}
Y.~LeCun, L.~Bottou, Y.~Bengio, and P.~Haffner, ``{Gradient-based learning applied to document recognition},'' \emph{Proc. IEEE}, vol.~86, no.~11, pp. 2278--2324, 1998.

\bibitem{mcmahan2017communication}
B.~McMahan, E.~Moore, D.~Ramage, S.~Hampson, and B.~A. y~Arcas, ``{Communication-efficient learning of deep networks from decentralized data},'' in \emph{Proc. Artif. Intell. Stat. (AISTATS)}.\hskip 1em plus 0.5em minus 0.4em\relax PMLR, 2017, pp. 1273--1282.

\end{thebibliography}

\end{document}